\documentclass[runningheads]{llncs}

\usepackage[T1]{fontenc}          

\usepackage{graphicx}             
\usepackage{float}                
\usepackage[section]{placeins}
\usepackage{placeins}  \usepackage{subcaption}            

\usepackage[skip=0.5ex]{caption}  
\usepackage{booktabs}             
\usepackage{multirow}             
\usepackage{adjustbox}            
\usepackage{tabularx}             

\usepackage{makecell}

\usepackage{courier}
\usepackage{listings}             
\lstdefinestyle{promptstyle}{
  backgroundcolor=\color{gray!10},
  basicstyle=\ttfamily\small,
  keywordstyle=\bfseries\color{blue},
  commentstyle=\color{gray},
  stringstyle=\color{orange},
  showstringspaces=false,
  frame=single,
  breaklines=true,
  columns=fullflexible,
  escapeinside={(*@}{@*)},
}
\usepackage[ruled,vlined,linesnumbered]{algorithm2e}

\usepackage{amsmath}
\usepackage{amsfonts}

\usepackage[x11names]{xcolor}
\definecolor{boxgray}{HTML}{F7F7F7}
\definecolor{titlebar}{HTML}{2E3B4E}
\usepackage[most]{tcolorbox}
\usepackage{tabularx}
\usepackage{tikz}
\usetikzlibrary{tikzmark,arrows.meta,calc}
\usepackage{enumitem}
\usepackage{mathtools}
\usepackage{xcolor}
\usepackage[
  colorlinks=true,
  linkcolor=black,
  citecolor=black,
  filecolor=black,
  urlcolor=blue
]{hyperref}

\tcbset{
  mybox/.style={
    enhanced,
    sharp corners,
    boxrule=0.6pt,
    colback=boxgray,
    colframe=black!50,
    colbacktitle=titlebar,
    coltitle=white,
    fonttitle=\sffamily\large,
    left=4pt,right=4pt,top=3pt,bottom=3pt,
    lower separated=false,
    boxsep=1pt,
    breakable,
  }
}

\usepackage{xspace}

\let\subparagraph\relax
\usepackage{titlesec}
\titlespacing*{\section}{0pt}{0.5em}{0.5\baselineskip}
\titlespacing*{\subsection}{0pt}{1.0ex}{0.5ex}
\titleformat*{\paragraph}{\normalfont\bfseries}
\titlespacing*{\paragraph}{0pt}{0.5ex}{0.5ex}

\usepackage{enumitem}

\newcommand{\eat}[1]{}
\newcommand{\stab}{\rule{0pt}{8pt}\\[-1.6ex]}

\newcommand{\bi}{\begin{itemize}}
\newcommand{\ei}{\end{itemize}}
        {\end{itemize}\vspace{-0.5ex}}

\newcommand{\etitle}[1]{\vspace{0.8ex}\noindent{\em #1}}
\newcommand{\eetitle}[1]{\vspace{0.8ex}\noindent{\underline{\em #1}}}

\newcommand{\be}{\begin{enumerate}}
\newcommand{\ee}{\end{enumerate}}
\newcommand{\beqn}{\begin{eqnarray*}}
\newcommand{\eeqn}{\end{eqnarray*}}

\newcommand{\stitle}[1]{\vspace{1.5ex}\noindent{\bf #1}}

\newcommand{\ie}{\emph{i.e.,}\xspace}
\newcommand{\eg}{\emph{e.g.,}\xspace}

\newcommand{\kw}[1]{{\ensuremath{\mathsf{#1}}}\xspace}

\newcommand{\I}{{\mathcal I}}
\newcommand{\G}{{\mathcal G}}
\newcommand{\Q}{{\mathcal Q}}
\newcommand{\C}{{\mathcal C}}
\newcommand{\D}{{\mathcal D}}
\newcommand{\T}{{\mathcal T}}
\newcommand{\M}{{\mathcal M}}

\renewcommand{\P}{{\mathcal P}}

\newcommand{\kroma}{\kw{KROMA}}
\newcommand{\llm}{\kw{LLM}}
\newcommand{\llms}{\kw{LLMs}}

 \definecolor{softgreen}{HTML}{6B8E23}
 
 \definecolor{darkyellow}{HTML}{B8860B}

\definecolor{assignblue}{HTML}{4682B4} 

\newenvironment{proofS}{
        \vspace{1ex}
        {\noindent\bf Proof sketch:\ }}{\eop\vspace{1ex}}

        \newcommand{\eop}{\hspace*{\fill}\mbox{$\Box$}}     
 
\usepackage{etoolbox}
\AtEndEnvironment{table}{\vspace{0.5ex}}

\newcounter{qnumber}
\usepackage{listings}

\begin{document}

\title{KROMA: Ontology Matching with Knowledge Retrieval and Large Language Models}

\author{Lam Nguyen \and Erika Barcelos \and Roger French \and Yinghui Wu}
\authorrunning{L. Nguyen et al.}
\institute{Case Western Reserve University, Cleveland OH 44106, USA \\
\email{\{ltn18,eib14,rxf131,yxw1650\}@case.edu}}
\titlerunning{\kroma}
\maketitle

\begin{abstract}
Ontology Matching (OM) is a cornerstone task of semantic interoperability, yet existing systems often rely on handcrafted rules or specialized models with limited adaptability. We present KROMA, a novel OM framework that harnesses Large Language Models (LLMs) within a Retrieval‑Augmented Generation (RAG) pipeline, to dynamically enrich the semantic context of OM tasks with structural, lexical, and definitional knowledge. To optimize both performance and efficiency, KROMA integrates a bisimilarity‑based concept matching and a lightweight ontology refinement step, which prune candidate concepts and substantially reduce the communication overhead 
from invoking LLMs. Through experiments on multiple benchmark datasets, we show that integrating knowledge retrieval with context-augmented LLMs significantly enhances ontology matching—outperforming both classic OM systems and cutting-edge LLM-based approaches—while keeping communication overhead comparable. Our study highlights the feasibility and benefit of the proposed optimization techniques (targeted 
knowledge retrieval, prompt enrichment, and ontology refinement) for ontology matching at scale. Our code and experimental dataset has been made available at: \url{https://github.com/lamng3/kroma}
\end{abstract}

\section{Introduction}
\label{sec-intro}


Ontologies have been routinely developed to unify and standardize knowledge representation to support 
data-driven applications. 
They 
allow researchers to harmonize terminologies and enhance knowledge and sharing in their fields. 
Ontologies can be classified according to their level of specificity, ranging from more abstract, general-purpose ontologies such as Basic Formal Ontology (BFO) \cite{Otte2022-OTTBBF} or DOLCE \cite{Borgo_Ferrario_Gangemi_Guarino_Masolo_Porello_Sanfilippo_Vieu_2022}  down to  more domain-oriented `mid-level'' ones,  
such as CheBi \cite{Degtyarenko_Hastings_Matos_Ennis_2009} in chemistry, Industrial Ontology Foundy (IOF) \cite{drobnjakovicIndustrialOntologiesFoundry}  or Common Core Ontology (CCO) \cite{Jensen_Colle_Kindya_More_Cox_Beverley_2024}. Domain ontologies are data-driven, task-specific ``low-level'' ontologies, containing concepts from domain-specific data, such as Materials Data Science Ontology (MDS-Onto)~\cite{Rajamohan_Bradley_Tran_Gordon_Caldwell_Mehdi_Ponon_Tran_Dernek_Kaltenbaugh_et_al}.
To achieve broad usability, ontologies need to be effectively aligned for better interoperability using ontology matching.  

Ontology matching 
has been studied to find correspondence between terms that are semantically equivalent~\cite{trojahn2022foundational}. It is a cornerstone task to ensure semantic interoperability among terms originated from different sources. Ontology matching methods can be categorized 
to rule-based or 
 structural-based (graph pattern or path-based) matching \cite{cruzStructuralAlignmentMethods2008}, matching with semantic similarity, machine learning-based approaches and hybrid methods. Linguistic (terminological) methods are often used for ontology matching tasks, ranging from simple string matching or embedding learning to advance counterparts based on natural language processing (NLP). 
Nevertheless, conventional rule- or structural-based methods are often restricted to 
certain use cases and hard to be generalized for new or unseen concepts. 
Learning-based approaches may on the other hand 
require expensive re-training process, for which abundant annotated or training data remains a luxury especially for \eg 
data-driven scientific research.



Meanwhile, the emerging Large Language Models (\llms) 
have demonstrated remarkable versatility for NL understanding. LLMs are trained on vast and diverse corpora, endowing them with an understanding of language nuances and contextual subtleties. Extensive training enables them to capture semantic relationships and abstract patterns that are critical for aligning concepts across different data sources. 
A missing yet desirable opportunity is to investigate whether and how \llms can be engaged   
to automate and improve ontology matching.

This paper introduces \textbf{\kroma}, a novel framework that exploits \llms to enhance 
 ontology matching. Unlike existing \llm-based 
 methods that typically ``outsource'' 
ontology matching to \llms using direct prompting (which may result in low confidence 
and risk of hallucination), 
\kroma maintains a set of conceptually 
similar groups that are co-determined by 
concept similarity and language models, both 
guided by their ``augmented context'' 
obtained via a runtime knowledge retrieval process. 
Moreover, the groups are further refined 
by a global ontological equivalence relation 
that incorporate structural equivalence. 

\stitle{Contributions}. Our main technical contributions are summarized below. 

\stab 
(1)  We propose a formulation of semantic equivalence relation in terms of a class of 
bisimilar equivalence relation,  
and formally define the ontology structure, 
called concept graph, to be maintained (\textbf{Sections~\ref{sec-pre} and~\ref{sec-problem}}). 
We justify our formulation by showing 
the existence of an ``optimal'' 
concept graph with minimality and uniqueness 
guarantee, subject to the bisimilar equivalence. 

\stab 
(2) We introduce \kroma, an 
ontology matching framework leveraging the power of \llms (\textbf{Section~\ref{sec-framework}}). 
  \kroma fine-tunes \llms with prompts over enriched semantic contexts. Such 
contextual information are obtained from the knowledge retrieval process, referencing high-quality, external knowledge resources. 

\stab 
(3) \kroma supports both offline and online 
matching, to ``cold-start'' from 
scratch, and to digest terms arriving from a 
stream of data, respectively. 
We introduce efficient algorithms 
to correctly construct and maintain the 
optimal concept graphs (\textbf{Section~\ref{sec-refine}}). 
(a) Offline refinement algorithm
performs a fast grouping process 
guided by the bisimilar equivalence property, blending the concept 
equivalence tests co-determined by 
semantic closeness and language models. (b) Online refinement algorithm effectively incrementalizes its offline counterpart with fast delay time for continuous concept streams. Both algorithms are in low polynomial time, with optimality guarantee on the computed concept graphs. 

\stab 
(4) Using benchmarking ontologies and knowledge graphs, 
we experimentally verify the effectiveness and 
efficiency of \kroma (\textbf{Section~\ref{sec-exp}}). We found that 
\kroma outperforms existing methods using large languge models by 10.95\% on average, and the optimization of utilizing knowledge retrieval and refinement processes improves \kroma's accuracy by 6.65\% and 2.68\%, respectively.

\stitle{Related Work}. We summarize the related work below.   

\etitle{Large Language Models}. Large language models (LLMs) have advanced NLP by enabling parallel processing and capturing complex dependencies \cite{seq2seq,transformers}, which have scaled from GPT-1’s 117M \cite{gpt1}, GPT-2’s 1.5B \cite{gpt2} to GPT-3’s 175B \cite{gpt3} and GPT-4’s 1.8T parameters \cite{gpt4}. Open-source models like Llama have grown to 405B~\cite{Llama2}, with Mistral Large (123B) \cite{mistrallarge2} and DeepSeek V3 (671B) \cite{deepseekv3} also emerging. Recent advances in specialized reasoning LLMs (RLLMs) such as OpenAI’s O1 and DeepSeek R1 have further propelled Long Chain‑of‑Thought reasoning—shifting from brief, linear Short CoT to deeper, iterative exploration, and yielded substantial gains in 
multidisciplinary tasks \cite{chen2025evaluatingo1likellmsunlocking,deepseekai2025deepseekr1incentivizingreasoningcapability,li2025surveyllmtesttimecompute,openai2024openaio1card,qin2024multilinguallargelanguagemodel,sun2024surveyreasoningfoundationmodels,kimiteam2025kimik15scalingreinforcement,Xu_2025}.

\etitle{Ontology Matching with LLMs}. Several methods have been developed 
to exploit LLMs for ontology matching. %
Early work focused on direct prompting LLMs for ontology matching. For example, \cite{peeters2023usingchatgptentitymatching} frame product matching as a yes/no query, and \cite{norouzi2023conversationalontologyalignmentchatgpt} feed entire source and target ontologies into ChatGPT—both achieving high recall on small OAEI conference‐track tasks but suffering from low precision.  Beyond these “direct‐prompt” approaches, state-of-the-art LLM‐OM systems fall into two main categories: (1) retrieval‐augmented pipelines, which first retrieve top‑\(k\) candidates via embedding‑based methods and then refine them with LLM prompts (e.g.\ LLM4OM leverages TF–IDF and SBERT retrievers across concept, concept‑parent, and concept‑children representations \cite{llm4om}, while MILA adds a prioritized depth‑first search step to confirm high‑confidence matches before any LLM invocation \cite{mila}); and (2) prompt‐engineering systems, which generate candidates via a high‑precision matcher or inverted index and then apply targeted prompt templates to LLMs in a single step (e.g.\ OLaLa embeds SBERT candidates into MELT’s prompting framework with both independent and multi‑choice formulations \cite{olala}, and LLMap uses binary yes/no prompts over concept labels plus structural context with Flan‑T5‑XXL or GPT‑3.5 \cite{llmap}).

\section{Ontologies and Ontology Matching}
\label{sec-pre}

\begin{table}[h]
\centering
\begin{tabular}{|>{\centering\arraybackslash}p{4.5cm}|p{7.5cm}|}
\hline
\textbf{Notation} & \textbf{Description} \\
\hline
$\mathcal{O} = (C, E)$ & ontology $\mathcal{O}$, $C$: set of concepts, $E$: set of relations \\
\hline
$|\mathcal{O}|$ & size of ontology $\mathcal{O}$; $|\mathcal{O}| = |C| + |E|$ \\
\hline
$\text{r}(c)$ & rank of concept node $c$ \\
\hline
$c.I$ & ground set of concept node $c$ \\
\hline
$\mathcal{O}_s = (C_s, E_s)$, $\mathcal{O}_t = (C_t, E_t)$ & source and target ontology, respectively \\
\hline
$R_\simeq$ & ontological equivalence relation \\
\hline
$\mathcal{C}$ & equivalence partition of concept set $C_s\cup C_t$ \\
\hline
$\mathcal{G}_O = (V_{\mathcal{O}}, E_{\mathcal{O}})$ & concept graph $\mathcal{G}_O$, $V_O$: set of nodes, $E_O$: set of edges \\
\hline
$\Delta\mathcal{G}$ & newly arrive edges; edge updates in $\mathcal{G}$ \\
\hline
$[c] \in V_O$ & an equivalence class in $\mathcal{C}$\\
\hline
$M$ & a language model \\
\hline
$\mathcal{M}$ & set of Large Language Model(s) \\
\hline
$q$ & a prompt query \\
\hline
$q(M)$ & a natural language answer from language model $M$ \\
\hline
$F(c, c')$ & concept similarity between two concepts $c$ and $c'$ \\
\hline
$q(c, c', \mathcal{M})$ & a natural language answer to $\mathcal{M}$ asking if $c \simeq c'$\\
\hline
$\alpha \in (0, 1]$ & threshold for asserting concept similarity \\
\hline
$W = (O_s, O_t, F, \M, \alpha)$ & configuration input for ontology matching \\
\hline
$z_c$ & embedding of concept $c$ \\
\hline
$\mathbb{S}$ & set of candidate concept pairs with similarity scores\\
\hline
$\mathbb{C}$ & top-k pairs with highest similarity scores \\
\hline
\end{tabular}
\caption{Summary of Notations.}
\label{tab:notation_summary}
\end{table}

\vspace{-1em}
\stitle{Ontologies}. An ontology 
$O$ is a pair $(C, E)$, where $C$ 
is a finite set of concept names, and 
$E\subset C\times C$ is a 
set of relations between the concepts. An ontology has a graph representation 
with a set of concept nodes 
$C$, and a set of edges 
$E$. In this paper, we consider ontologies as directed acyclic graphs (DAGs).

In addition, each concept node (or simply ``node'') 
$c$ in $O$ carries the following auxiliary structure. 
(1) The {\em rank} of a node $c\in C$ is defined 
as: (a) $r(c)$ = 0 if $c$ has no child, 
otherwise, (b) $r(c)$ = $\max(r(c')+1)$ 
for any child $c'$ of $c$ in $O$. 
(2) A {\em ground set} $c.\I$, 
refers to a set of auxilary entities 
from \eg external ontologies or 
knowledge bases, that can be 
validated to belong to the 
concept $c$. 

\stitle{Ontology Matching}. 
Given a source ontology $O_s$ = $(C_s, E_s)$ and  
a target ontology $O_t$ = $(C_t, E_t)$, an  
{\em ontological equivalence relation} 
$R_\simeq\subseteq C_s\times C_t$  
is defined as an equivalence relation that 
contains pairs of nodes $(c,c')$ that are considered to be ``semantically equivalent''. The relation $R_\simeq$ induces an equivalence partition $\C$ of the concept set $C_s\cup C_t$, such that each partition is an equivalence class that contains a set of pairwise equivalent concepts in $C_s\cup C_t$. 

Consistently, we define 
a {\em concept graph} $\G_O$ = $(V_O, E_O)$ as a 
DAG with a finite set of nodes $V_O$, where each 
node $[c]\in V_O$ is an equivalence class 
in $\C$, and there exists an edge 
$([c],[c'])\in E_O$ if and only if 
there exists an edge $(c, c')$ in $E_s$ or 
in $E_t$. A concept graph $\G_O$ can be a 
multigraph: there may exist multiple edges 
of different relation names between two 
equivalent classes. 

Given $O_s$ and $O_t$, the task of ontology matching is to 
formulate $R_\simeq$ and $\C$ induced by 
$R_\simeq$ and compute the concept graph $\G_O$ accordingly.  

\begin{figure}[tb!]
    \centering
    \includegraphics[width=0.9\linewidth]{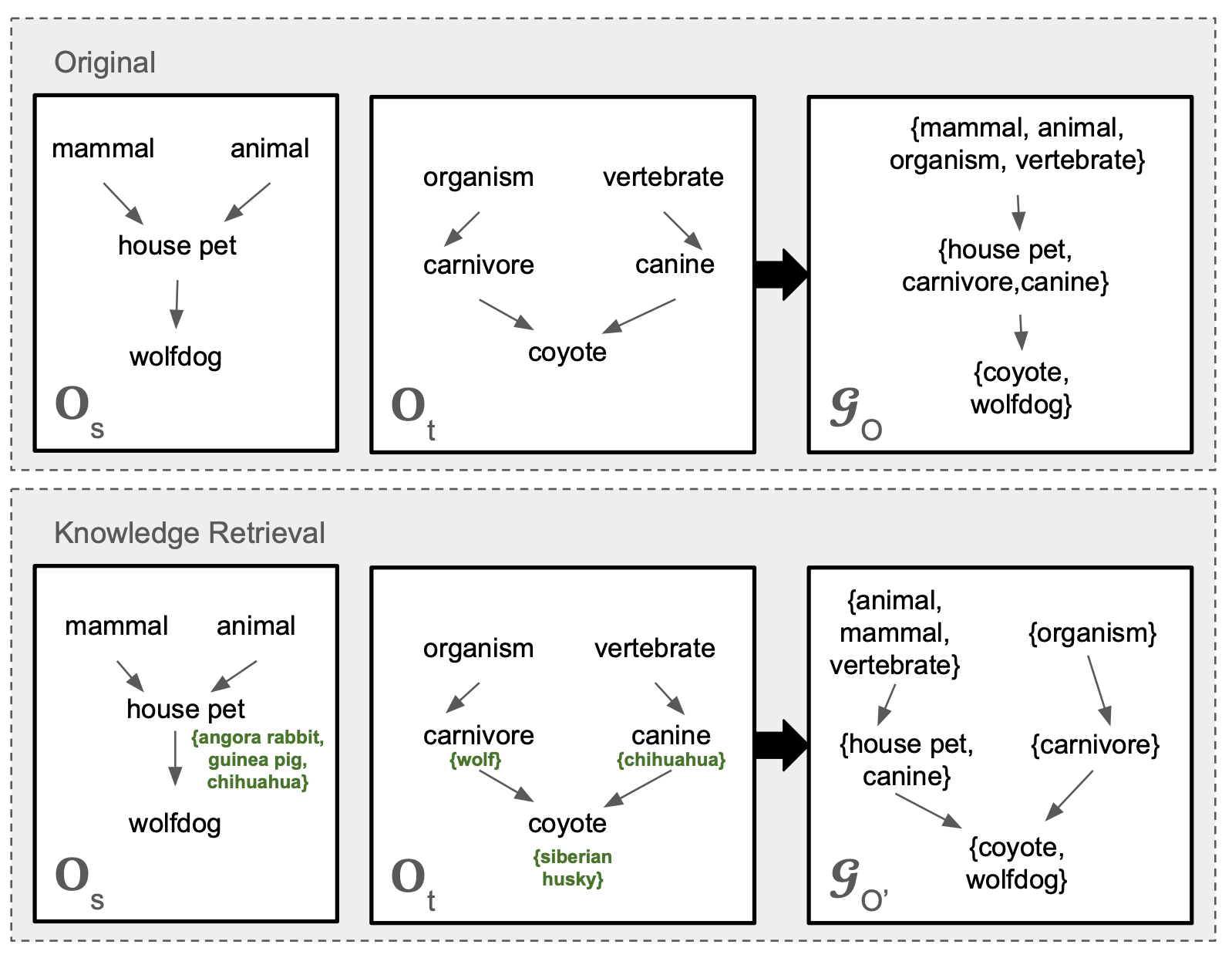}
    \caption{Ontologies with ground sets, Ontology Matching and Concept Graphs}
    \label{fig:example1}
\end{figure}

\begin{example}
\label{exa-onto}
Figure~\ref{fig:example1} depicts two ontologies $O_s$ and $O_t$ as DAGs with 9 concept nodes. An equivalence relation 
may suggest that ``mammal'', ``animal'', ``organism'' 
and ``vertebrate'' are pairwise similar; and similarly for the sets 
\{``house pet'', ``carnivora'' and ``canine''\}, 
and \{``wolfdog, ``coyote''\}''. 
This induces a concept graph $\G_o$ as a result of ontology matching, with three equivalence classes. 
\end{example}

\vspace{-1ex}
\stitle{Language Models for Ontology Matching}. 
A language model $M$ 
takes as input a prompt query $q$, and generate  
an answer $q(M)$, typically an NL statement, 
for downstream processing. Large language models 
(\llms) are foundation
models that can effectively learn from a handful of in‑context examples, included in a prompt query $q$, that demonstrate input–output distribution \cite{wei2022emergentabilitieslargelanguage}. 

\etitle{Prompt query}. A prompt query 
$q$ is in a form of NL statements that specifies 
(1) a task definition $\T$ with input and output 
statement; (2) a set of in-context examples $\D$ 
with annotated data; (3) a statement of query 
context $Q$, which describe auxiliary query semantics; and 
optionally (4) specification on output 
format, and (5) a self-evaluation of 
answer quality such as confidence. 
An evaluation of a prompt query $q$ 
invokes an \llm $M$ to infer a query result $q(M)$. 

We specify a prompt query $q$ and a large language model $M$ for ontology matching. A prompt query $q$ is in the form of 
$q(c, c')$, which asks ``are $c$ and $c'$ 
semantically equivalent?'' An \llm $M$ 
can be queried by $q(c, c')$ and acts as a Boolean ``oracle'' with ``yes/no'' answer. An \llm is 
{\em deterministic}, if it always generate 
a same answer for the same prompt query. 
We consider deterministic \llms, as in practice, such \llms are desired for consistent
and robust performance. 

\section{Ontology Matching with LLMs}
\label{sec-problem}

In this section, we provide a pragmatic 
characterization for the ontological equivalence relation 
$R_\simeq$. We then formulate 
the ontology matching problem. 

\subsection{Semantic Equivalence: A Characterization} 

\stitle{Concept similarity}. 
A variety of methods have been proposed to 
determine whether two {\em concepts} are semantically equivalent~\cite{chandrasekaran2021evolution}. \kroma by default uses a Boolean  
function $F$ defined by a weighted combination 
of a semantic closeness metric $\kw{sim}$\footnote{We 
adopt similarity metrics that satisfy transitivity, \ie 
if concept c is similar to c', and c' is similar to c'', 
then c is similar to c''. This is to ensure 
transitivity of ontology equivalence, and is practical for 
representative concept similarity measures.} and the result from a set of \llms $\M$ (see Section~\ref{sec-framework}).
\[
F(c, c') = \gamma \kw{sim}(c,c') + (1-\gamma) q(c, c', \M)
\]
where $q$ is a prompt query that 
specifies the context of concept equivalence 
for \llms, and $\gamma$ be a configurable parameter. 
\kroma supports a built-in library of semantic 
similarity functions \kw{sim}, including 
(a) string similarity, 
feature and information  measure~\cite{pirro2010feature}, 
or normalized distances (NGDs)~\cite{jiang2014semantic}; and 
(b) a variety of \llms such as GPT-4o Mini \cite{gpt4}, LLaMA-3.3 \cite{llama3}, and Qwen-2.5 \cite{qwen2025qwen25technicalreport}.

\stitle{Ontological Bisimilarity}. 
We next provide a specification of the 
ontological equivalence relation, notably, 
{\em ontological bisimilarity}, 
denoted by the same symbol $R_\simeq$ for simplicity. 
Given a source ontology $O_s$ = $(C_s, E_s)$, 
and a target ontology $O_t$ = $(C_t, E_t)$, 
we say a pair of nodes $c_s\in C_s$ and $c_t\in C_t$ 
are {\em ontologically bisimilar}, denoted as $(c_s, c_t)\in R_\simeq$, 
if and only if there exists a non-empty binary relation $R_\simeq$, 
such that: (1)  $c_s$ and $c_t$ are conceptually similar, \ie $F(c_s,c_t)\geq \alpha$, for a threshold $\alpha$; (2) for every edge $(c_s', c_s)\in E_s$, 
there exists an edge $(c_t', c_t)\in E_t$, 
such that $(c_s', c_t')\in R_\simeq$,    
and vice versa; and
(3) for every edge $(c_s, c_s'')\in E_s$, 
there exists an edge $(c_t, c_t'')\in E_t$, 
such that $(c_s'', c_t'')\in R_\simeq$. 

\begin{lemma}
\label{lm-equiv}
The ontological bisimilar relation $R_\simeq$ is an equivalence relation. 
\end{lemma}

We can prove the above results by verifying that 
$R_\simeq$ is reflexive, symmetric, and transitive 
over the concept set $C_s\cup C_t$, ensured by the 
transitivity of concept similarity and by  
definition. Observe that two entities that are conceptually similar may \textit{not} be ontologically bisimilar. On the other hand, two ontologically bisimilar entities must be conceptually similar, following the definition. 

\subsection{Problem Statement}

We now formulate our ontological matching problem. 
Given a {\em configuration}  
$W$ = $(O_s, O_t, F, \M, \alpha)$ that specifies as input a source ontology $O_s$, a target 
ontology $O_t$, a Boolean function $F$ and 
threshold $\alpha\in (0,1]$ that determines concept similarity, and 
a set of \llms $\M$, the problem becomes 
computing a smallest concept graph $\G_O$ 
induced by the ontologically bisimilar equivalence $R_\simeq$. 

We can justify the above characterization 
by proving that there exists an ``optimal'', invariant solution encoded by a concept graph $\G_O$. To arrive at this, we provide a \textit{minimality} and \textit{uniqueness} guarantee on $\G_O$ as below. 

\begin{lemma}
\label{lm-uniq}
Given a configuration $W$ and semantic equivalence  specified by the ontological bisimilar relation $R_\simeq$ property, 
there is a unique smallest concept graph $\G_O$ that captures all semantically equivalent nodes 
in terms of $R_\simeq$.  
\end{lemma}

\begin{proofS}
We show that the above result holds by 
verifying the following. (1) There is a unique, 
maximum ontological bisimilar relation $R_\simeq$ 
for a given configuration $W$ = $(O_s, O_t, F, \M, \alpha)$, 
where any $\llm$ in $\M$ is a deterministic 
model for the same prompt query $q$ generated 
consistently from $W$. This readily follows from 
Lemma~\ref{lm-equiv}, which verifies that $R_\simeq$ 
is an equivalence relation. 
(2) Let the union of $O_s$ 
and $O_t$ be a graph $O_{st}$ = $\{C_s\cup C_t, E_s\cup E_t\}$. 
By setting $\G_O$ as the quotient graph 
induced by the largest ontological bisimilar 
relation $R_\simeq$ over $O_{st}$, 
$\G_O$ contains the smallest number of 
equivalent classes (nodes) and edges. 
This can be verified by proof by contradiction. 
(3) The uniqueness of the solution can be verified 
by showing that $R_\simeq$ induces only 
one unique partition $\C$ and results in 
a concept graph $\G_O$ up to graph isomorphism. 
In other words, for any two possible 
concept graphs induced by $R_\simeq$, 
they are isomorphic to each other. 
\end{proofS}

The above analysis suggests that for a configuration $W$, 
it is desirable to compute such an optimal concept graph as an invariant, stable result with guarantees on sizes and uniqueness on topological structures. 
We next introduce \kroma and efficient algorithms to compute the aforementioned
optimal concept graphs. 
\section{KROMA Framework}
\label{sec-framework}

\begin{figure}[tb!]
    \centering
    \includegraphics[width=0.98\linewidth]{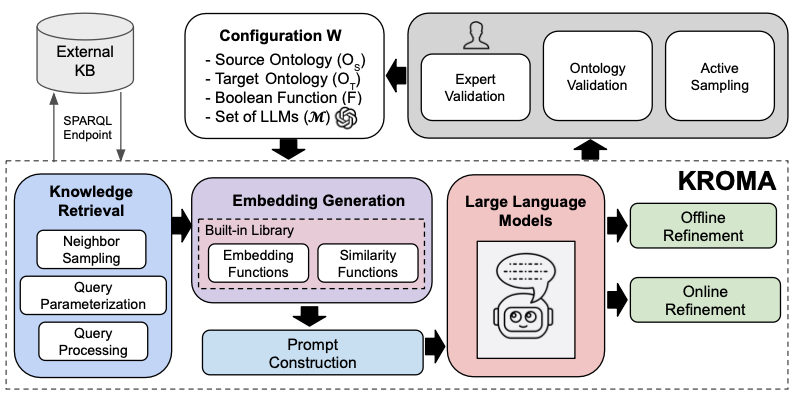}
    \caption{\kroma Framework Overview: Major Components and Dataflow}
    \label{fig:framework}
\end{figure}

\subsection{Framework Overview}
Given a {\em configuration}  
$W$ = $(O_s, O_t, F, \M)$ where $\M$ is a set of pre-trained large language models $\mathcal{M}$, \kroma has the following major functional modules that enables a robust and effective multi-session ontology matching process. 

\eetitle{Concept Graph Initialization}
Upon receiving two ontologies $O_s$ = $(C_s,E_s)$ and $O_t$ = $(C_t,E_t)$ as the inputs, \kroma initializes the concept graph $\mathcal{G}_O$ = $(V, E)$ with $V = C_S \cup C_T$ and $E = E_S \cup E_T$. For each concept $c\in V$, \kroma then globally computes the structural $rank$ $r(c)$ as described in Section~\ref{sec-pre}.

\eetitle{Knowledge Retrieval.}
To assemble a rich, yet compact, context for each concept \(c\in V\), we perform: 
(1) {\em Neighborhood Sampling}: traverse up to its two hops in $\mathcal{G}_O$ to collect parents, children, and ``sibling'' concepts of $c$, creating a node induced subgraph centered at $c$; 
(2) {\em Subgraph Parameterization}: Sample and substitute 
constant values from the subgraph with variables to create SPARQL queries $S_q$; and 
(3) {\em Ground Set Curation}: Apply the generated SPARQL queries $S_q$ onto external knowledge bases to augment the concept's ground set $c.\I$ with auxiliary information (\eg relevant entities, definition, labels, synonyms, etc.). 

\eetitle{Embedding Generation.} In this phase, \kroma leverages a built-in library of semantic similarity functions $\kw{sim}(\cdot,\cdot)$ and embedding functions $\kw{embd}(\cdot)$. For each concept $c\in V$, \kroma computes the joint textual and structural embedding: 
$$
z_c = \alpha\;\kw{embd}_{graph}(c) + (1-\alpha)\;\kw{embd}_{text},
$$
where $\kw{embd}_{graph}$ (e.g. node2vec \cite{grover2016node2vecscalablefeaturelearning}) captures the $c$'s topology information and $\kw{embd}_{text}$ (e.g. SciBERT \cite{beltagy2019scibertpretrainedlanguagemodel}) encodes $c$'s textual context. After obtaining the necessary embeddings ($z_c$ $\forall c\in V$), \kroma then computes pairwise concept similarity between source and target ontologies using the \kw{sim} functions:
$$
\mathbb{S} = \{ (c_s, c_t, score_{s,t}) \mid c_s\in C_S, c_t \in C_T, score_{s,t} = \kw{sim(z_{c_s}, z_{c_t})} \}
$$
From \(\mathbb{S}\), we select the top-\(k\) pairs with the highest similarity scores (i.e.\ in descending order of \(score_{s,t}\)), yielding the best candidate list \(\mathbb{C}\) to ask \llms.

\begin{example}
\label{exa-ontosim}
We revisit Example 1. (1) A knowledge retrieval for 
 node ``house pet'' samples 
 its neighbors and issues 
a set of ``star-shaped'' 
SPARQL queries centered at 
``house pet'' to query an underlying  
knowledge graph $KG$. 
This enriches its ground set 
with a majority of herbivore or omnivorous 
pets that are not ``carnivore''. 
Similarly, the ground 
set of ``carnivore'' is enriched
by ``wolf'', unlikely a house pet. 
Despite ``coyote'' and ``wolfdog'' (house pet) alone are less similar, 
the ground set of ``coyote''
turns out to be a set of canine pets \eg ``husky'' that are ``coyote-like'', hence similar with ``wolfdog''. 
(2) The embedding generation phase incorporates 
 enriched ground sets and generate embeddings accordingly, which scores that distinguishes   
``house pet'' from 
``carnivore'' due to embedding difference, and assert ``coyote'' and ``wolfdog'' to be conceptually similar, allowing the candidate pair to be ``double checked'' by \llms in the following phase.  
\end{example}

\vspace{-1ex}
\eetitle{Prompt Querying LLMs.} 
For each candidate pair \((c_s,c_t)\in\mathbb{C}\), \kroma generates an NL prompt that includes: (1) Task description $\mathcal{T}$ (e.g. ``\textit{Given two ontology concepts and their metadata, decide if they are related or not.}''), 
(2) In‐context examples $\mathcal{D}$ containing both positive and negative matches, (3) Query context for \(c_s\) and \(c_t\) including their ground sets, (4) Output format and confidence (e.g., ``Answer Yes or No, and provide a confidence score between 0 and 10.''). \kroma then calls (a set of) \llms $\mathcal{M}$ to obtain a matching decision with confidence.  Low‐confidence or conflicting outputs are routed to validation module. 

\begin{figure}[tb!]
    \centering
    \includegraphics[width=0.95\linewidth]{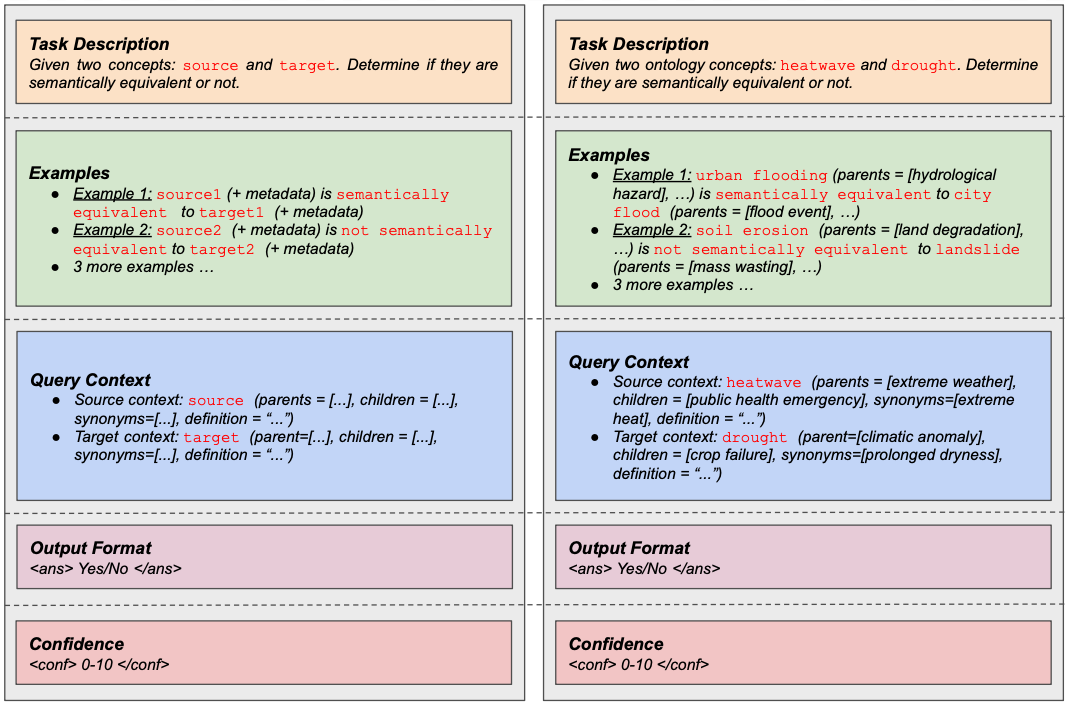}
    \caption{\kroma Prompt Query Template and Query Example}
    \label{fig:prompt_template}
\end{figure}

A query template and a generated example 
is illustrated in Figure~\ref{fig:prompt_template}. 

\eetitle{Ontology Refinement \& Expert Validation.}
\kroma next invokes a refinement process, \(\mathsf{OfflineRefine}\) (Algorithm\,\ref{alg:offline_refine}), to ``cold-start'' the construction of $\G_O$, or \(\mathsf{OnlineRefine}\) (Algorithm 2), to incrementally refine $\mathcal{G}_O$ for any unseen, newly arrived 
concept nodes or edges.  Any pair of nodes whose structural ranks remain in ``conflict'' is routed into a set of queries for expert validation (see Algorithm 2, Section~\ref{sec-framework}); once approved, are integrated back into $\G_O$. An active sampling 
strategy is applied, to select pairs of nodes that have low confidence from \llms 
for expert validation, with an aim to minimalize the manual effort needed. 

\begin{example} 
Continuing with Example 2, 
as ``golden retriever'' and 
``coyote'' are asserted by 
the function $F$ that combines 
the descision of semantic similarity 
function $\kw{sim}$ and \llms, 
an equivalence class is created 
in $\G_O'$. As 
``house pet'' and ``carnivore'' 
has quite different embedding 
considering the features 
from themselves and their 
ground sets, ``carnivore'' is 
separated from the  
group of ``house pet''. 
This suggests further that 
``organism'' now has a different context 
that distinguish it from the group 
\{mammal, animal, vertebrate\}, 
by the definition of bisimilarity 
equivalence. This leads to a finer-grained  
concept graph $\G_O'$.  
\end{example} 

\section{Ontology Refinement}
\label{sec-refine}

\begin{algorithm}[tb!]
\caption{Offline Refinement}\label{alg:offline_refine}
\KwIn{Source ontology $O_S=(C_S,E_S)$, target ontology $O_T=(C_T,E_T)$} 
\KwOut{Concept graph $\G_O$.}
set $V \leftarrow C_S\cup C_T$; set $E \leftarrow E_S\cup E_T$\; 
Initialize concept graph $\G_O=(V,E)$\;
\ForEach{$c\in V$}{
  compute rank $r(c)$ as in Section 2\;
}
$\rho\leftarrow\max_{c\in V}r(c)$\;
\For{$i\leftarrow 0$ \KwTo $\rho$}{
  $B_i\leftarrow\{c:r(c)=i\}$\;
}
$P\leftarrow\{B_0,\dots,B_\rho\}$\;
\For{$i\leftarrow 0$ \KwTo $\rho$}{
  $D_i\leftarrow\{X\in P:X\subseteq B_i\}$\;
  \ForEach{$X\in D_i$}{
    $G\leftarrow\textsf{collapse}(G,X)$\;
  }
  \ForEach{$c\in B_i$}{
    \ForEach{$C\in P$ with $C\subseteq\bigcup_{j>i}B_j$}{
      Split $C$ into $C_1,C_2$ by adjacency to $c$\;
      $P\leftarrow(P\setminus\{C\})\cup\{C_1,C_2\}$\;
    }
  }
}
\Return $\G_O$\;
\end{algorithm}

We next describe our ontology refinement algorithms. 
\kroma supports ontology refinement in two modes. 
The offline mode assumes that the source ontology 
$O_s$ and the target ontology $O_t$ are known, 
and performs a batch processing to compute 
the concept graph $\G_O$ from scratch. 
The online refinement incrementally 
maintains $\G_O$ upon 
a sequence of (unseen) triples (edges) 
from external resources. 

The offline refinement algorithm is outlined 
as Algorithm 1. (1) It starts by initializing 
$\G_O$ (lines 1-6) as the union of $O_s$ and $O_t$, 
followed by computing the node ranks. At each 
rank, it initializes a ``bucket'' $B_i$ 
(as a single node set) that simply include all the 
concept nodes at rank $i$ (lines 7-8), 
and initialize a partition $\P$ with all the 
buckets (line 9). It then follows a 
``bottom-up'' process to refine 
the buckets iteratively, by checking 
if two concepts $c$ and $c'$ in a same bucket $B_i$ 
are concept similar (as asserted by \llm and 
embedding similarity), and have 
all the neighbors that satisfy the requirement 
of ontological bisimilar relation by 
definition (lines 10-13). If not, 
a procedure \kw{collapse} is invoked, 
to (1) split the bucket $B_i$ into 
three fragments: $B^1_i$ = $B_i\setminus\{c,c'
\}$, $B^2_i$ = $\{c\}$, and $B^3_i$=$\{c'\}$, 
followed by a ``merge'' check to test 
if $c$ and $c'$ can be merged to $B^1_i$; 
and (2) propagate this change to further 
``split-merge'' operators to affected 
buckets at higher ranks (lines 14-17). 
This process continues until no change 
can be made.

\eetitle{Correctness}. Algorithm 1 correctly 
terminates at obtaining a maximum bisimilar ontological 
equivalence relation $R_\simeq$, with two 
variants. (1) As ontologies are DAGs, it suffices to 
perform a one-pass, bottom-up splitting of 
equivalence classes following the topological ranks; 
(2) the \kw{collapse} operator 
ensures the ``mergable'' cases to reduce 
unnecessary buckets whenever a new bucket is 
separated. This process simulates 
the correct computation of maximum bisimulation 
relation in Kripke structures (a DAG)~\cite{dovier2001fast}, 
optimized for deriving ontology matching 
determined by \llm-based concept similarity 
and bisimilar equivalence. 

\eetitle{Time Cost.} The initialization of concept graph 
$\G_O$ is in $O(|O_s|+|O_t|)$. Here $|O_s|$=$|V_s|+|E_s|$; 
and $|O_t|$ is defined similarly. 
The iterative collapse (lines 10-17)
takes in $O(|O_s|+|O_t|)$ time as 
the number of buckets (resp. edges) is at most $|C_s|+|C_T|$ 
(resp. $|E_s|$+$|E_t|$). 
The overall cost is in $O(|O_s|+|O_t|)$.  

\eetitle{Overall Cost}. We consider the cost of the entire workflow of \kroma. 
(1) The knowledge checking takes $O((|C_s|+|C_T|)|KG|)$ time, 
where $|KG|$ refers to the size of the external 
ontology or knowledge graph that is referred to by the knowledge retrieval via 
\eg SPARQL access. Note here we consider SPARQL queries 
with ``star'' patterns, hence the cost of query processing (to curate ground sets)
is in quadratic time. (2) The total cost of \llm 
inference is in $O(|C_s||C_t|T_I)$, 
for a worst case that any pair of nodes in $C_s\times C_t$ 
are concept similar in terms of $\sim$. Here $T_I$ 
is the unit cost of processing a prompt query. 
Putting these together, the total cost 
is in $O(|C_s||C_t|T_I+(|C_s|+|C_t|)|KG|+(|O_s|+|O_t|))$ 
time.

\stitle{Online Refinement}.
We next outline the online matching process. In this setting, 
\kroma receives new ontology components as an (infinite) sequence of triples (edges), and incrementally maintain a concept graph $\G_O$ by 
processing the sequence input in small batched 
updates $\Delta\G$. 
For each newly arrived concept (node) $c$ in $\Delta \G$, \kroma conducts knowledge retrieval to curate $c.\I$; and 
consult \llms to decide if $c$ is concept similar 
to any node in $\G_O$. It then 
invokes online refinement algorithm 
to enforce the ontological bisimilar equivalence. 
  
The algorithm (with pseudoscope reported in~\cite{full}) 
first updates the buckets in $\G_O$ by 
incorporating the nodes from $\Delta \G$ that are verified to be 
concept similar, as well as their ranks. It then incrementally  
update the buckets to maintain the bisimilar equivalence consistently via a ``bottom-up'' split-merge process as in Algorithm 1 (lines 6-15). 
Due to the unpredictability of the ``unseen'' 
concepts, the online 
refinement defers the processing of two 
``inconsistent'' cases for experts' 
validation: (1) 
When a concept $c$ is determined to be 
concept similar by function $\sim$ but 
not \llms with high confidence; or 
(2) whenever for a new edge $(c, c')\in\Delta \G$, 
$(c, [c_1])\in R_\simeq$, 
$(c', [c_2])\in R_\simeq$, 
yet $r(c_1)<r(c_2)$ in $\G_O$. 
Both require domain experts' 
feedback to resolve.
These cases are cached into 
a query set $\Q$ to be 
further resolved in the 
validation phase (see Section~\ref{sec-framework}).
We cache these cases into a query set by using an auxiliary data structure (\eg using a priority queue ranked by \llm confidence scores) to effectively manage their processing.

\eetitle{Analysis.} The correctness of 
online refinement carries over from 
its offline counterpart, and that it 
correctly incrementalize the split-merge 
operator for each newly arrived 
edges. 
For time cost, for each batch, 
it takes a delay time to update 
$\G_O$ in $O(|\G_O| + |\Delta G|\log |\Delta G| + |\Delta G|\log|V_O|)$ time. 
This result verifies that 
online refinement is able to 
response faster than 
offline maintenance 
that recomputes  
the concept graph from scratch. 
We present detailed 
analysis in~\cite{full}. 

\eat{
The Online Ontology Refinement procedure runs in near-linear time in the size of the compressed graph and the update batch.  Recomputing all node ranks on $\mathcal{G}_r$ costs $O(N + M)$ (with $N=|V_r|$, $M=|E_r|$), and sorting the $\Delta$ new edges by rank takes $O(\Delta\log\Delta)$.  Each insertion is then processed via union-find splits and merges in $O\bigl((M + \Delta)\,\alpha(N)\bigr)$ overall—since every edge participates in at most one split or merge and each such operation is $O(\alpha(N))$, where $\alpha$ is the inverse Ackermann function.  Finally, dispatching $Q$ deferred pairs to the human verifier incurs an extra $O(Q)$.  Altogether, the time complexity is
$$
O\bigl(N + M + \Delta\log\Delta + (M + \Delta)\,\alpha(N) + Q\bigr)\,,
$$
with space overhead $O(N + M + \Delta + Q)$.
}


\section{Experimental Study}
\label{sec-exp}

We investigated the following research questions. 
\textbf{[RQ1]}: \textit{How well \kroma improves state-of-the-art baselines with different \llms?}
\textbf{[RQ2]}: \textit{How can knowledge 
retrieval and ontology refinement enhance matching performance?} and 
\textbf{[RQ3]}: \textit{What are the impact of using alternative \llm reasoning strategies?}

\subsection{Experimental Setting}


\stitle{Benchmark Datasets}. We selected five tracks from the OAEI campaign~\cite{qiang2024oaei}, covering various domain tracks. For each track, we selected two representative ontologies as a source ontology $O_s$ and a target ontology $O_t$. 
The selected tracks include Anatomy \cite{anatomy} (Mouse-Human), Bio-LLM \cite{llmap} (NCIT-DOID), CommonKG (CKG) \cite{commonkg} (Nell-DBpedia, YAGO-Wikidata), BioDiv \cite{biodiv} (ENVO-SWEET), and MSE \cite{mse} (MI-MatOnto). To ensure a fair and comprehensive evaluation of KROMA, we adopted the standard benchmarks from the Ontology Alignment Evaluation Initiative (OAEI), enabling direct comparison with prior LLM-based methods. Despite the high cost of LLM inference, we tested KROMA across five diverse tracks to demonstrate its robustness and effectiveness across domains.

\stitle{\llms selection.} To underscore \kroma’s ability to achieve strong matching performance even with smaller or lower-performance LLMs, we selected models with relatively modest Chatbot Arena MMLU scores \cite{chiang2024chatbotarenaopenplatform}: Gemma-2B (51.3\%) and Llama-3.2-3B (63.4\%), compared to the baseline systems Flan-T5-XXL (55.1\%) and MPT- 7B (60.1\%). We have selected a diverse set of \llms, ranging from ultra-lightweight to large-scale—to demonstrate \kroma's compatibility with models that can be deployed on modest hardware without sacrificing matching quality. Our core evaluations use DeepSeek-R1-Distill-Qwen-1.5B \cite{qwen1.5} (1.5B), GPT-4o-mini \cite{gpt4} (8B), and Llama-3.3 \cite{llama3} (70B), each chosen in a variant smaller than those employed by prior OM-LLM baselines. To further benchmark our framework, we include Gemma-2B \cite{google2024gemma2b} (2B), Llama-3.2-3B \cite{meta2024llama32} (3B), Mistral-7B \cite{mistral2023mistral7b} (7B), and Llama-2-7B \cite{meta2023llama2} (7B) in our ablation studies. To run inference on the aforementioned models, we used TogetherAI and OpenAI APIs, treating them as off-the-shelf inference services on hosted, pretrained models. We are not performing any fine-tuning or weight updates to the models.


\eetitle{Confidence Calibration.} Our selection of \llms is justified by a calibration test with their 
confidence over ground truth answers. The self-evaluated confidence by \llms align well with performance: over 80\% of correct matches fall in the top confidence bins (9–10), while fewer than 5\% of errors are reported. Gemma-2B shows almost no errors above confidence 8, and both Llama-3.2-3B and Llama-3.3-70B maintain $\geq95\%$ precision at confidence thresholds of 9 or greater. We thus choose a confidence threshold to be 8.5 for all \llms to accept their output.

\stitle{Test sets generation}. 
Following 
~\cite{llm4om,llmap}, for each dataset, 
we arbitrarily designate one concept as the ``source'' for sampling and its target counterpart. We remark that the source and target roles are interchangeable w.l.o.g given our theoretical analysis, algorithms and test results. We randomly sample 20 matched concept pairs from the ground truth mappings. For each source ontology concept, we select an additional 24 unmatched target ontology concepts based on their cosine similarity scores, thereby creating a total of 25 candidate mappings (including the ground truth mapping). Finally, we randomly choose 20 source concepts that lack a corresponding target concept in the ground truth and generate 25 candidate mappings for each. Each subset consists of 20 source ontology concepts with a match and 20 without matches, with each concept paired with 25 candidate mappings, totaling 1000 concept pairs per configuration. 

Concepts and entities not selected as test sets are treated as external knowledge base for \textbf{knowledge retrieval}. All models operate with SciBERT \cite{beltagy2019scibertpretrainedlanguagemodel} for \textbf{Embedding Generation}, leveraging its 
strength in scientific data embedding. 

\stitle{Baselines.} 
Our evaluation considers \textbf{four} state-of-the-art  
\llm-based ontology matching methods: LLM4OM \cite{llm4om}, MILA \cite{mila}, OLaLa \cite{olala}, and LLMap \cite{llmap}. For RQ2, we also developed KROMA‑NR, which skips the knowledge retrieval process, and KROMA‑NB, which disables the use of bisimilarity‑based clustering, allowing only the use of concept similarity to determine concept clusters. 

\eetitle{Naming Convention.}
Each configuration uses a pattern \texttt{[Method][Optional Suffix][LLM Version]}, where the initials (K, M, O, L) specify the method (KROMA, MILA, OLaLa, LLM4OM). Suffixes ``NKR'' and ``NR'' denote ``no knowledge retrieval'' and ``no ontology refinement'', respectively, and the trailing version number (e.g., 3.3, 2.0, 4mini) specifies the underlying 
LLM release. 

\subsection{Experimental Results}

\stitle{Exp-1: Effectiveness (RQ1)}. In this set of tests, 
we evaluate the performance of \kroma compared with baselines, 
and the impact of factors such as test sizes. 

\begin{table}[!ht]
  \centering
  \caption{\kroma Performance vs. Baselines.}
  \setlength{\tabcolsep}{2pt}
  \renewcommand{\arraystretch}{0.8}
  \begin{tabular}{l c c c c c c}
    \toprule
    Model
      & MH
      & ND
      & NDB
      & YW
      & ES
      & MM \\
    \midrule
    KL3.3      & 94.94 & 98.63 & 97.08 & 95.54 & –     & 61.25 \\
    KNR3.3     & 91.25 & 95.01 & 94.26 & 91.98 & –     & 59.95 \\
    KNB3.3     & 86.59 & 90.91 & 93.58 & 89.74 & –     & 55.00 \\
    KL3.1      & 94.50 & 98.24 & –     & –     & 91.43 & –     \\
    KL2.0      & 93.24 & –     & 96.02 & –     & 85.06 & –     \\
    KG2        & –     & 85.53 & –     & –     & –     & –     \\
    KM7        & –     & –     & –     & –     & 92.98 & –     \\
    ML3.1      & 92.20 & 94.80 & –     & –     & 83.70 & 32.97 \\
    OL2.0      & 90.20 & –     & 96.00 & –     & 51.10 & –     \\
    L4G3.5     & 89.11 & 83.01 & 94.26 & –     & –     & –     \\
    K4mini     & –     & –     & –     & –     & 93.18 & –     \\
    LLFT       & –     & 72.10 & –     & –     & –     & –     \\
    L4L2       & –     & –     & –     & 92.19 & –     & –     \\
    L4M7       & –     & –     & –     & –     & 55.09 & –     \\
    L4MPT      & –     & –     & –     & –     & –     & 32.97 \\
    \bottomrule
  \end{tabular}
  \label{tab:kroma_baseline_overall_accuracy}
\end{table}

\eetitle{\kroma vs. Baselines: Overall Performance}. 
Across all six datasets in Table \ref{tab:kroma_baseline_overall_accuracy} (abbreviated by their first capitalized letters), the full \kroma configuration (KL3.3) achieves the highest $F_1$ on every task, substantially outperforming competing baselines.  For Mouse–Human, KL3.3 delivers 94.94 $F_1$, eclipsing MILA’s best (ML3.1) at 92.20, OLaLa (OL2.0) at 90.20, and LLM4OM (L4G3.5) at 89.11.  On NCIT–DOID, KL3.3 reaches 98.63 versus 94.80 for MILA, 83.01 for LLM4OM, and 72.10 for Flan-T5-XXL.  Similar gaps appear on Nell–DBpedia (97.08 vs.\ 96.00/94.26), YAGO–Wikidata (95.54 vs.\ 92.19/93.33), ENVO–SWEET (93.18 vs.\ 92.98/85.06), and MI–MatOnto (61.25 vs.\ 59.05/32.97). These cumulative results verify 
the effectiveness of \kroma over representative benchmark datasets. More details (\eg Precision 
and Recall) are reported in~\cite{full}.

\eetitle{Impact of different \llms}. 
Across six ontology-matching tracks, KROMA equipped with Qwen2.5-1.5B 
outperforms the best existing baseline on five out of the six datasets (see Table \ref{tab:kroma_with_different_llms}).  In the Anatomy's Mouse–Human track, Qwen2.5-1.5B achieves $F_1=83.58$ (versus 92.20 for the OM-LLM baseline), while GPT-4o-mini reaches $F_1=91.96$.  On NCIT–DOID both models surpass the baseline with $F_1$ of 97.44 and 97.56 (baseline: 94.80), and similar gains appear on Nell–DBpedia (95.02, 95.67 vs. 96.00), YAGO–Wikidata (95.45, 95.44 vs. 92.19), and MI–MatOnto (61.25, 60.88 vs. 32.97).  Only on ENVO–SWEET does the smallest model dip below the baseline (79.80 vs. 83.70), whereas GPT-4o-mini (93.18) and Llama-3.3-70B (91.95) still lead. These results show that KROMA (with both knowledge retrieval and ontology refinement) delivers strong performance even on relatively ``small'' LLMs, and scales further with larger \llms.  

\begin{table}[!ht]
  \centering
  \caption{\kroma Performance with different \llms.}
  \setlength{\tabcolsep}{2pt}
  \renewcommand{\arraystretch}{0.8}
  \begin{tabular}{l c c c}
    \toprule
    Dataset
      & Qwen2.5
      & GPT-4o-mini
      & Llama-3.3 \\
    \midrule
    Mouse–Human    & 83.6  & 92.0  & \textbf{94.9} \\
    NCIT–DOID      & 97.4  & 97.6  & \textbf{98.6} \\
    Nell–DBpedia   & 95.0  & 95.7  & \textbf{97.1} \\
    YAGO–Wikidata  & 95.5  & 95.4  & \textbf{95.5} \\
    ENVO–SWEET     & 79.8  & \textbf{93.2}  & 92.0  \\
    MI–MatOnto     & 61.3  & 60.9  & \textbf{62.2} \\
    \bottomrule
  \end{tabular}
  \label{tab:kroma_with_different_llms}
\end{table}

\eetitle{Impact of Test sizes}. We next report the impact of ontology sizes to the 
performance of \kroma in performance (detailed results reported in~\cite{full}). 
For each dataset, we varied test sizes from 200 (xsmall) to 1,000 (full) pairs. \kroma's performance is in general insensitive to the change of 
test sizes. For example, its $F_1$ stays within a 1.90 variance on NCIT–DOID 
and a 1.10-point range on YAGO–Wikidata. 
This verifies the robustness and effectiveness of \kroma in maintaining desirable performance for large-scale ontology matching tasks. 

\stitle{Exp-2: Ablation Analysis (RQ2)}.
In this test, 
we perform ablation analysis, comparing \kroma with 
its two variants, \kroma-NKR and \kroma-NR, removing 
knowledge retrieval and ontology refinement, respectively. 

\begin{table}[ht]
  \caption{\kroma performance under different scenarios: with/without  Ontology Refinement (Left); and with/without Knowledge Retrieval (Right).}
  \centering
  \scriptsize         
  \begin{subtable}[t]{0.48\textwidth}
    \centering
    \begin{tabular}{l l c c c}
      \toprule
      Dataset & Model & \textbf{P} & \textbf{R} & \textbf{F$_1$} \\
      \midrule
      \multirow{2}{*}{Mouse–Human}
        & KL3.3   & 90.78 & 99.50 & \textbf{94.94} \\
        & KNR3.3  & 92.34 & 90.16 & 91.25 \\
      \midrule
      \multirow{2}{*}{NCIT-DOID}
        & KL3.3   & 97.59 & 99.69 & \textbf{98.63} \\
        & KNR3.3  & 93.20 & 96.90 & 95.01 \\
      \midrule
      \multirow{3}{*}{Nell-DBpedia}
        & KL3.3   & 94.32 &100.00 & \textbf{97.08} \\
        & KNR3.3  & 97.46 & 91.27 & 94.26 \\
      \midrule
      \multirow{2}{*}{YAGO-Wikidata}
        & KL3.3   & 91.80 & 99.60 & 95.54 \\
        & KNR3.3  & 93.50 & 90.50 & 91.98 \\
      \midrule
      \multirow{2}{*}{ENVO-SWEET}
        & K4mini  & 87.23 &100.00 & \textbf{93.18} \\
        & KNR4mini& 86.90 & 98.00 & 92.12 \\
      \midrule
      \multirow{2}{*}{MI-MatOnto}
        & KL3.3   & 45.74 & 92.69 & \textbf{61.25} \\
        & KNR3.3  & 44.80 & 91.40 & 59.95 \\
      \bottomrule
    \end{tabular}
    \label{tab:impact_of_ontology_refinement}
  \end{subtable}%
  \hfill
  \begin{subtable}[t]{0.48\textwidth}
    \centering
    \begin{tabular}{l l c c c}
      \toprule
      Dataset & Model & \textbf{P} & \textbf{R} & \textbf{F$_1$} \\
      \midrule
      \multirow{2}{*}{Mouse–Human}
        & KL3.3   & 90.78 & 99.50 & \textbf{94.94} \\
        & KNKR3.3 &100.00 & 76.36 & 86.59 \\
      \midrule
      \multirow{2}{*}{NCIT-DOID}
        & KL3.3   & 97.59 & 99.69 & \textbf{98.63} \\
        & KNKR3.3 & 83.33 &100.00 & 90.91 \\
      \midrule
      \multirow{2}{*}{Nell-DBpedia}
        & KL3.3   & 94.32 &100.00 & \textbf{97.08} \\
        & KNKR3.3 & 91.17 & 96.12 & 93.58 \\
      \midrule
      \multirow{2}{*}{YAGO-Wikidata}
        & KL3.3   & 91.80 & 99.60 & 95.54 \\
        & KNKR3.3 & 90.50 & 89.00 & 89.74 \\
      \midrule
      \multirow{2}{*}{ENVO-SWEET}
        & K4mini   & 87.23 &100.00 & \textbf{93.18} \\
        & KNKR4mini& 82.00 & 88.00 & 84.89 \\
      \midrule
      \multirow{2}{*}{MI-MatOnto}
        & KL3.3   & 45.74 & 92.69 & \textbf{61.25} \\
        & KNKR3.3 & 42.00 & 85.00 & 55.00 \\
      \bottomrule
    \end{tabular}
    \label{tab:impact_of_knowledge_retrieval}
  \end{subtable}
  \label{tab:impact_of_or_and_kr}
\end{table}

\eetitle{Impact of Ontology Refinement}. 
Table~\ref{tab:impact_of_or_and_kr} (Left) shows that by incorporating Ontology Refinement (\kroma vs. \kroma-NR), \kroma improves $F_1$ score across 6 datasets: Mouse–Human (+3.69), NCIT–DOID (+3.62), Nell–DBpedia (+2.82), YAGO–Wikidata (+3.56), ENVO–SWEET (+1.06), MI–MatOnto (+1.30). By pruning false candidate pairs using offline and online strategies on the concept graph, \kroma is able to retain true semantical connections between concepts.

\eetitle{Impact of Knowledge Retrieval.} 
Table \ref{tab:impact_of_or_and_kr} (Right) shows that ablating Knowledge Retrieval (\kroma-KNR vs. \kroma) results in significant $F_1$ drop across all six benchmark datasets: Mouse–Human (–8.35), NCIT–DOID (–7.72), Nell–DBpedia (–3.50), YAGO–Wikidata (–5.80), ENVO–SWEET (–8.29), MI–MatOnto (–6.25). By enriching concepts with external, useful semantic context that are overlooked by other baselines, knowledge retrieval helps improving the performance.

\eat{
\begin{table}[tb!]
  \centering
  \setlength{\tabcolsep}{4pt}               
  \caption{\kroma Reduction in \llms API usage.}
  \begin{tabular}{l c c c}
    \toprule
    Dataset      & Qwen2.5 & GPT-4o-mini & Llama-3.3 \\ 
    \midrule
    Mouse-Human           &  3.06\%               &  2.88\%              &  6.76\%               \\ 
    NCIT-DOID             & 54.48\%               & 54.63\%              & 54.48\%               \\ 
    Nell-DBpedia          & 26.09\%               & 25.76\%              & 26.09\%               \\ 
    YAGO-Wikidata         & 21.86\%               & 21.93\%              & 22.01\%               \\ 
    ENVO-SWEET            &  6.99\%               &  6.82\%              &  6.99\%               \\ 
    MI-MatOnto            & 43.04\%               & 41.66\%              & 41.81\%               \\ 
    \midrule
    \textbf{Overall}          & \textbf{25.92\%}      & \textbf{25.61\%}     & \textbf{26.36\%}      \\ 
    \bottomrule
  \end{tabular}
  \label{tab:api_reduce}
\end{table}
}
\stitle{Exp-3: Alternative \llm Reasoning Strategies}. 
We evaluated \kroma's performance with two alternative \llm reasoning strategies: ``Debating'' and ``Deep reasoning'' to 
further understand their impact to ontology matching.

\begin{table*}[!htb]
  \caption{\kroma~Performance with/without  ``Debating'' (Left); and with/without ``Deep reasoning'' (Right).}
  \scriptsize
  \setlength{\tabcolsep}{1pt}        
  \renewcommand{\arraystretch}{0.8}   
  
  \begin{tabularx}{\textwidth}{@{} 
      >{\raggedright\arraybackslash}X 
      >{\centering\arraybackslash}X @{}}
    
    \begin{tabular*}{0.49\textwidth}{@{\extracolsep{\fill}} l l c c c @{}}
      \toprule
      Dataset        & Model   & P   & R   & F$_1$   \\ 
      \midrule
      \multirow{4}{*}{Mouse–Human}
        & D2A3R   & 100 & 70  & 82.35   \\
        & D2A5R   & 100 & 74  & \textbf{85.06}   \\
        & D4A3R   & 100 & 66  & 79.52   \\
        & D4A5R   & 100 & 68  & 80.95   \\
      \addlinespace
      \multirow{4}{*}{Nell–DBpedia}
        & D2A3R   & 97.83 &  90 & 93.75   \\
        & D2A5R   & 95.91 &  94 & \textbf{94.95}   \\
        & D4A3R   & 95.55 &  86 & 90.53   \\
        & D4A5R   & 93.88 &  92 & 92.93   \\
      \addlinespace
      \multirow{4}{*}{YAGO–Wikidata}
        & D2A3R   & 100   &  92 & 95.83   \\
        & D2A5R   & 100   & 96.9& \textbf{98.41}   \\
        & D4A3R   & 100   &  86 & 92.47   \\
        & D4A5R   & 100   &  90 & 94.73   \\
      \bottomrule
    \end{tabular*}
    &
    \begin{tabular*}{0.49\textwidth}{@{\extracolsep{\fill}} l l c c c @{}}
      \toprule
      Dataset        & Model               & P    & R    & F$_1$   \\ 
      \midrule
      \multirow{2}{*}{NCIT–DOID}
        & \makecell{L3.3Short} & 97.59 & 99.69 & \textbf{98.63} \\
        & \makecell{L3.3Long}  & 97.66 & 96.31 & 96.98        \\
      \midrule
      \multirow{2}{*}{Nell–DBpedia}
        & \makecell{L3.3Short} & 94.32 &100.00 & \textbf{97.08} \\
        & \makecell{L3.3Long}  & 94.67 & 95.48 & 95.07        \\
      \midrule
      \multirow{2}{*}{YAGO–Wikidata}
        & \makecell{L3.3Short} & 91.80 & 99.60 & \textbf{95.54} \\
        & \makecell{L3.3Long}  & 92.31 & 94.80 & 93.53        \\
      \bottomrule
    \end{tabular*}
  \end{tabularx}

  \label{tab:alternative_llm_reasoning_strategies}
\end{table*}

\eetitle{Can ``Debating'' help?}
We implemented an LLM Debate ensemble \cite{llmdebate}, where multiple agents propose alignments with their chain-of-thought and then iteratively critique one another. To bound context size, we drop 50\% of historic turns and keep only each agent’s latest reply. Due to its expense, we tested this on Mouse-Human, Nell-DBpedia, and YAGO-Wikidata using four permutation of configurations: 2 or 4 agents over 3 or 5 debate rounds, denoted as \texttt{D[Number Of Agent]A[Number Of Round]R}. From Table \ref{tab:alternative_llm_reasoning_strategies} (Left), on Mouse–Human dataset, extending rounds in the 2-agent setup raised F1 from 82.35 to 85.06, whereas adding agents degraded performance (4 agents: 79.52 at 3 rounds, 80.95 at 5). Similar trends follow for Nell-DBpedia and YAGO-Wikidata. This interestingly indicates that ``longer debates'' help small ensembles converge to accurate matches, but larger groups introduce too much conflicting reasoning. In contrast, a single-agent, single-round KROMA pass attains F1 = 90.78, underscoring that a well-tuned solo model remains the most efficient and reliable choice.

\eetitle{Can ``Deep reasoning'' help?}
Building on DeepSeek-R1’s “Aha Moment” \cite{deepseekai2025deepseekr1incentivizingreasoningcapability}, we extended KROMA to include a forced long chain-of-thought for self-revision. Noting that vanilla DeepSeek-R1 often emits empty ``\texttt{<think>\textbackslash n}'' tags (i.e. $``\texttt{<think>$  $\textbackslash n$ $\textbackslash n$ $</think>}$''), we altered our prompt so every response must start with ``\texttt{<think>\textbackslash n}'', ensuring the model spells out its reasoning. We then compared standard Llama-3.3 (70B) (``short'' reasoning, noted as L3.3Short) to DeepSeek-R1-Distill-Llama-3.3 (70B) (``long'' reasoning, noted as L3.3Long), finding that the added "\texttt{<think>\textbackslash n}" tag inflated inputs by 20\% and outputs by 600\% but delivered only tiny precision gains (e.g., NCIT–DOID: 97.59\% → 97.66\%) while slashing recall (99.69\% → 96.31\%), dropping F1 by 1.65–2.01 points, based on Table \ref{tab:alternative_llm_reasoning_strategies} (Right). This suggests that verbose self-reflection may improve transparency but consumes crucial context and can overly filter valid matches—especially in binary tasks, where lengthy chains of thought have been shown to harm recall, which has been consistently observed in \cite{wei2023chainofthoughtpromptingelicitsreasoning}.

\eetitle{Can ``Active Sampling'' help?}
In the semi-supervised experiments on the Bio-LLM NCIT-DOID benchmark, KROMA paired with Llama-3.3-70B matches or exceeds the performance of MILA + Llama-3.1-70B.  Under an active-learning regime with few-shot demonstrations, our method scores 1.90 F1 points higher than MILA and surpasses the previous state of the art by over 6 points.  As Table \ref{tab:active_learning} shows, combining bisimilarity-guided refinement with selective querying delivers a substantial boost in \kroma's alignment performance.

\vspace{-1em}
\begin{table}[!htb]
  \centering
  \scriptsize                               
  \setlength{\tabcolsep}{4pt}               
  \caption{Active Learning on NCIT-DOID.}
  \begin{tabular}{l l c c c}
    \toprule
    \textbf{Dataset}     & \textbf{Model}            & \textbf{$P$} & \textbf{$R$} & \textbf{$F_1$} \\ 
    \midrule
    \multirow{2}{*}{NCIT-DOID}
      & KROMA + Llama-3.1-70B   & 97.85 & 98.15 & \textbf{98.00} \\
      & MILA  + Llama-3.1-70B   & 96.70 & 92.80 & 94.67          \\
    \bottomrule
  \end{tabular}
  \label{tab:active_learning}
\end{table}

\section{Conclusion}
We have presented \kroma, an ontology matching framework that exploits the semantic capabilities of \llms within a bisimilar-based ontology refinement process. We show that \kroma computes a provably unique minimized structure that captures 
semantic equivalence 
relations, with efficient algorithms 
that can significantly reduces LLMs 
communication overhead, while achieving state-of-the-art performance across multiple OAEI benchmarks. Our evaluation has verified that LLMs, when guided by context and optimized prompting, can rival or surpass much larger models in performance. 
A future topic is to extend 
\kroma with more 
\llm reasoning strategies 
and ontology engineering tasks. 

\stitle{Supplemental Material Statement}. Our code and experimental dataset has been made available at \url{https://github.com/lamng3/kroma}, including  
an extended version of the paper~\cite{full}, providing further details for \kroma. 

\stitle{Acknowledgements.} This material is based upon research in the Materials Data Science for Stockpile Stewardship Center of Excellence (MDS3-COE), and supported by the Department of Energy's National Nuclear Security Administration under Award Number(s) DE-NA0004104. This material is based upon work supported by the U.S. National Science Foundation, Office of Advanced Cyberinfrastructure (OAC), Major Research Instrumentation, under Award Number 2117439. This work made use of the High Performance Computing Resource in the Core Facility for Advanced Research Computing at CWRU.


\clearpage
\bibliographystyle{splncs04}
\bibliography{references}
\end{document}